# K-nearest Neighbour and Dynamic Time Warping for Online Signature Verification

Mohammad Saleem and BenceKovari

Department of Automation and Applied Informatics,
Budapest University of Technology and Economics, Budapest, Hungary

*ABSTRACT*

*Online signatures are one of the most commonly used biometrics. Several verification systems and public databases were presented in this field. This paper presents a combination of k-nearest neighbor and dynamic time warping algorithms as a verification system using the recently published DeepSignDB database. Our algorithm was applied on both finger and stylus input signatures which represent both office and mobile scenarios. The system was first tested on the development set of the database. It achieved an error rate of 6.04% for the stylus input signatures, 5.20% for the finger input signatures, and 6.00% for a combination of both types. The system was also applied to the evaluation set of the database and achieved very promising results, especially for finger input signatures.*

## 1. Introduction

Biometrics are used for authentication and identification purposes; signatures are widely used in bank checks, documents, payments, and many other fields. Signatures are classified into online and offline signatures based on the input methods. In online signatures, special devices are used to acquire the signatures, such as tablets and digital pens. These devices can capture several features like position, pressure, azimuth, and altitude as a function of time (see table 1). Online signatures have more features than offline signatures, where signatures are acquired using a regular pen and paper, then scanned and processed as an image. Thus, it is harder to forge an online signature compared to an offline signature. Nowadays, online signatures are used more frequently than before due to digital devices' development and the need for quick actions and smart methods to keep up withthe digital evolution.

Data acquisition, preprocessing, feature extraction, and verification are the main stages of any online signature verification system. These stages together form the process of classifying a signature as genuine or forged. Several methods and algorithms can be applied for each step. The effect of these different algorithms on the accuracy of the verification system varies for other systems.

Many databases are publicly available for researchers, such as the database of Signature Verification Competition2004 (SVC2004) [1], the Spanish Ministry of Science and Technology database (MCYT-100) [2], the Dutch and Chinese subsets of the Signature Verification

Competition 2011 database (SigComp'11) [3], BiosecureID [4][5], and the German database of the Signature Verification Competition 2015 (SigComp'15) [6]. In this work, a recently published database is used, the DeepSignDB [7]. These databases varyby the signees number, signature number, type of forgery, the input device used, and some other features.

The same signee provides very similar (yet not exactly the same) signatures. These signatures might vary by size, position, pressure, or any other feature. To reduce these inner-class differences, several preprocessing methods can be applied. Scaling and translation algorithms, where the signature is scaled and shifted to a specific range of points, are commonly used algorithms in data preprocessing. Some other methods can be applied, such as zero pressure removal, rotation, or resampling.

The last step of the verification is the classification phase, where several similarity measurements and classification algorithms can be applied to decide whetherthe questioned signature is genuine or forged. Then the system accuracy is evaluated using specific evaluation methods. All the previous steps will be discussed in detail throughout the paper.

In the following section, the related work from the literatureis briefly presented. Afterward, we describe our work methodology and present the experimental results. Finally, the results are evaluated, and the paper is concluded.

## 2. RELATED WORK

Dynamic time warping (DTW) is widely used for verification and similarity measurement. DTW finds the minimum distance between two-time series, which may vary in length [8]. It is one of the most commonly usedalgorithms measuring the similarities of time series. It has also shownpromising results in the field of online signature verification.

K-nearest neighbor (k-NN) algorithm is applied to calculate upper and lower thresholds of the proposed algorithms, which are then used for signature classification. k-NN is a one-class classification algorithm and was previously used in some verification systems [9] [10].

DTW and k-NN were applied in the state of the art of online signature verification field, but not together. Feng et al. proposed a warping technique for DTW [11]. Faundez-Zanuy proposed a method using a combination of vector-quantization and DTW [12], and Parziale et al. used stability-modulated DTW [13].

There are several published signature verification competition that compares different system on the same database such as the signature verification competition 2004 (SVC2004) [1], the international conference on document analysis and recognition ICDAR competitions (ICDAR 2009 [14]), (ICDAR 2013 [15]) and (ICDAR 2015 [6]), the Signature Competition 2011 (SigComp2011 [3]). These signature verification competitionsprovide fair comparisons between the verification systems applied on the same database under similar circumstances.

## 3. METHODOLOGY

DeepSignDB database is used in this work. It is a combination of the most commonly used databases in the field. More than 70,000 signatures were acquired in this database from 1526 signees [7]; see figure 1. It contains both stylus and finger inputs using eight different devices. The DeepSignDB is divided into two sets.

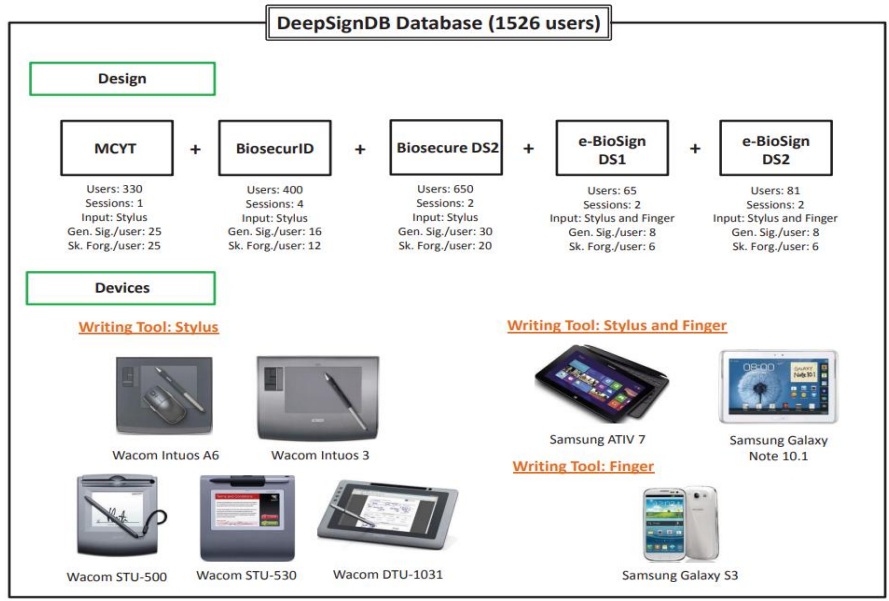

Figure 1. DeepSignDB database [7].

The development set is used for training purposes, and the evaluation set, which helps researchers to evaluate the efficiency and accuracy of their proposed systems. The development and evaluation sets represent 70% and 30% of the database, respectively.

Preprocessing phase is very important to enhance the accuracy of the similarity measurement and reduce the inner class effect. The signature points were filtered in this work by removing the points with zero or shallow pressure values except for the finger input signatures, as theyhave zero pressure values already for all their points.

Previously, we studied the effect of some preprocessing methods on verification accuracy [16]. Scaling and translation have shown a strong impact on accuracy; thus, both were applied to all the signatures to reduce the inner-class differences that occurred since the same signee might provide very similar signatures but still with different scaling and starting points. Thus, the following algorithms were applied to scale the signature to the [0,1] range and shift the center of gravity to the origin.

$$\hat{x}(i) = x_{newMin} + \frac{x(i) - x_{oldMin}}{x_{oldMax} - x_{oldMin}} * (x_{newMax} - x_{newMin}) \quad (1)$$

$$\hat{x}(i) = x(i) - \mu_x \quad (2)$$

Where $x_{oldMin}$ and $x_{oldMin}$ are the new min-max points range, and $\mu_x$ is the points mean.

Although many features can be extracted from online signatures, the combination of the horizontal position (X), vertical position (Y), and pressure (P)are used for the classification purpose, see Table 1. All the preprocessing and similarity measurements were applied to the XYP combination. It showed more accurate results compared to the individuallyused features.

Table 1. Feature's combination

| # | Feature |
|---|---|
| 1 | x-coordinate: X |
| 2 | y-coordinate: Y |
| 3 | Pressure: P |
| Combination | **XYP** |

The proposed system uses DTW and k-NN algorithms in the classification phase. DTW is used as a distance measurement between the signatures. The k-NN algorithm is used to select the reference signatures and calculate upper and lower threshold, which plays a significant part in calculating the prediction of the tested signature. The k-NN algorithm is applying using the following formula:

$$d(S, S_{nn}) < \theta \frac{1}{K} \sum_{k=1}^{K} d(S_{nn}, S_{knn}) \quad (3)$$

where $d$ is the distance between the questioned signature ($S$) and its nearest neighbor ($S_{nn}$), $S_{knn}$ represents the $k$-nearest neighbor, and $\theta$ is a threshold used for the classification calculations.

In the classification phase, the distance $d$ between the questioned signature and the reference signature is calculated using DTW. The distance is used to calculate the prediction for the questioned signature $P_q$ using a calculated forgery threshold $F_{th}$, genuine threshold $G_{th}$ and a scaling parameter $s$ as follows:

$$P_q = \frac{s*F_{th} - d_s}{s*F_{th} - G_{th}} \quad (4)$$

Where $P_q$ is the prediction value, $s$ is the scale, $F_{th}$ and $G_{th}$ are the forged and genuine thresholds, respectively.

The prediction values are between zero and one, where zero represents a genuine signature, while one represents a forged signature. Furthermore, a threshold is assigned to classify the signature as genuine or forged based on its prediction value. Both false acceptance rate (FAR) and false rejection rate (FRR) were considered in the accuracy evaluation. Several thresholds can be applied, and the best result is chosen using the equal error rate (EER) where FAR and FRR cross.

## 4. EXPERIMENTAL RESULTS

As mentioned in the previous sections, DeepSignDB is divided into a development set and evaluation set. It contains both mobile and office scenarios, random and skilled forgery. In the development phase, the comparisons were provided using two strategies, 1vs1 where only one signature is available as a reference, and 4vs1, where four reference signatures are available. In the evaluation phase, only 1vs1 comparisons were available. The proposed system showed strong performance, especially when using 4vs1 comparisons. For the development set of the DeepSignDB, the results achieved were as following:

Computer Science & Information Technology (CS & IT) 165

- **Task1**

In Task1, only stylus input signatures were available. It represents the office scenarios. Our system achieved 6.04% EER (see figure 2).

- **Task2**

In Task2, only mobile scenario signatures using finger input were considered. The system achieved an EER of 5.20%, see figure 3.

- **Task3**

In Task3, both mobile and office scenario signatures were considered. The system achieved an EER equal to 6.00%, see figure4.

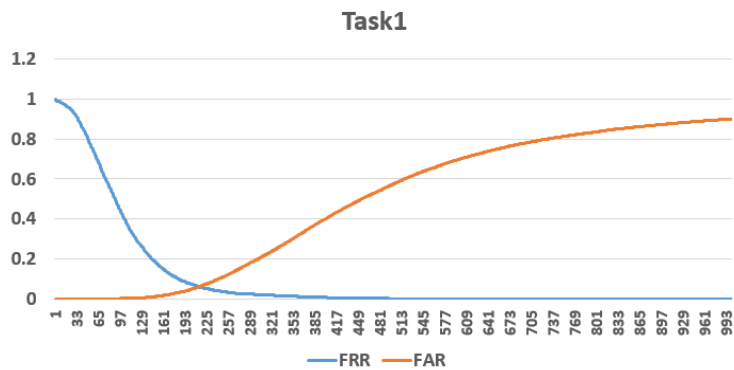

Figure 2. Task1 results.

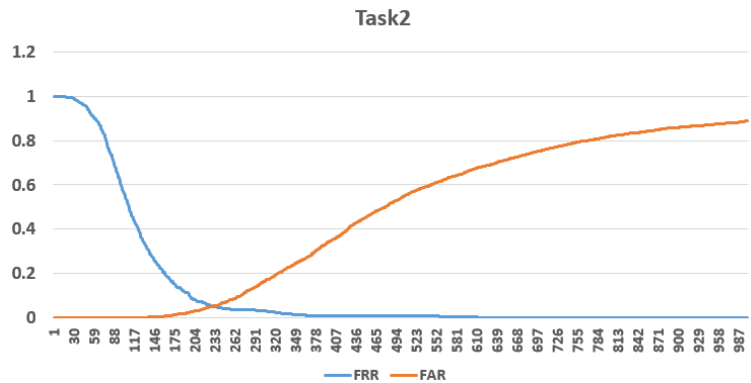

Figure 3. Task2 results.

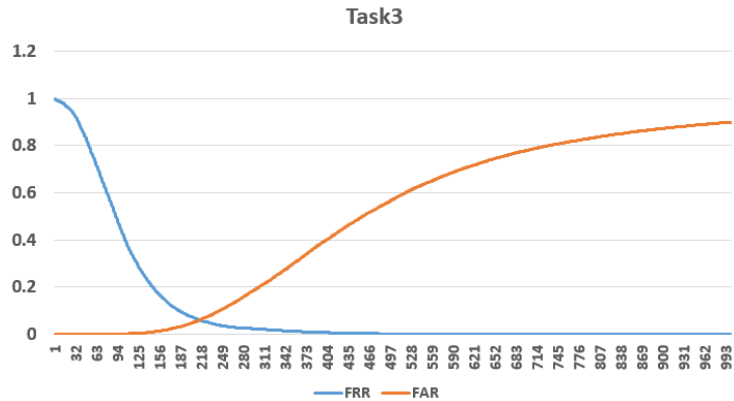

Figure 4. Task3 results.

• **Evaluation set**

For the evaluation set, the EER achieved was 13.28%; for this set, only 1vs1 scenarios were considered. This result is promising, and we believe it can be improved, which is currently under process.

The DeepSignDB database was published recently for the competition. Thus, there are not many published results using it. Table 2 shows a comparison with some other verification systems using different databases.

Table 2. A comparison between verification system for finger input signatures

| Study | Databse | EER |
|---|---|---|
| Tolosana et al. (2021) [7] | DeepSignDB | 13.8% |
| Lai and Jin (2018) [17] | Mobisig | 10.9% |
| Li et al. (2019) [18] | Mobisig | 16.1% |
| **Proposed** | **DeepSignDB** | **13.28%** |

## 5. CONCLUSION

A combination of the k-nearest neighbour and dynamic time warping algorithms is presented in this work as an online signature verification system. DeepSignDB was used for system development and evaluation, a combination of several databases with different input methods. The database is divided into development and evaluation sets where both office and mobile scenarios of signatures were used. The system achieved 6.04% EER when using stylus input signatures, 5.20% for finger input signatures, and 6.00% when using a combination of both types. The system also achieved 13.28% EER on the evaluation set of the database for finger input signatures. These promising results also showed that the system could be improved to adopt more scenarios and achieve higher accuracy.

**ACKNOWLEDGMENTS**

The work presented in this paper has been carried out in the frame of project no. 2019-1.1.1-PIACI-KFI-2019-00263, which has been implemented with the support provided from the National Research, Development and Innovation Fund of Hungary, financed under the 2019-1.1. funding scheme.

REFERENCES

[1] D.-Y. Yeung, H. Chang, Y. Xiong, S. George, R. Kashi, T. Matsumoto, and G. Rigoll, "Svc2004: First international signature verification competition," in International conference on biometric authentication, pp. 16–22, Springer, 2004.

[2] J. Ortega-Garcia, J. Fierrez-Aguilar, D. Simon, J. Gonzalez, M. Faundez-Zanuy, V. Espinosa, A. Satue, I. Hernaez, J.-J. Igarza, C. Vivaracho, et al., "MCYT baseline corpus: a bimodal biometric database," IEE Proceedings-Vision, Image and Signal Processing, vol. 150, no. 6, pp. 395–401, 2003.

[3] M. Liwicki, M. I. Malik, C. E. Van Den Heuvel, X. Chen, C. Berger, R. Stoel, M. Blumenstein, and B. Found, "Signature verification competition for online and offline skilled forgeries (sigcomp2011)," in 2011 International Conference on Document Analysis and Recognition, pp. 1480–1484, IEEE, 2011.

[4] J. Fierrez, J. Galbally, J. Ortega-Garcia, M. R. Freire, F. Alonso-Fernandez, D. Ramos, D. T. Toledano, J. Gonzalez-Rodriguez, J. A. Siguenza, J. Garrido-Salas, et al., "BiosecurID: a multimodal biometric database," Pattern Analysis and Applications, vol. 13, no. 2, pp. 235–246, 2010.

[5] J. Ortega-Garcia, J. Fierrez, F. Alonso-Fernandez, J. Galbally, M. R. Freire, J. Gonzalez-Rodriguez, C. Garcia-Mateo, J.-L. Alba-Castro, E. Gonzalez-Agulla, E. Otero-Muras, et al., "The multiscenario multi environment biosecure multimodal database (BMDB)," IEEE Transactions on Pattern Analysis and Machine Intelligence, vol. 32, no. 6, pp. 1097–1111, 2009.

[6] M. I. Malik, S. Ahmed, A. Marcelli, U. Pal, M. Blumenstein, L. Alewijns, and M. Liwicki, "Icdar2015 competition on signature verification and writer identification for on-and off-line skilled forgeries (sigwicomp2015)," in 2015 13th International Conference on Document Analysis and Recognition (IC-DAR), pp. 1186–1190, IEEE, 2015.

[7] R. Tolosana, R. Vera-Rodriguez, J. Fierrez, and J. Ortega-Garcia, "Deep-sign: Deep online signature verification," IEEE Transactions on Biometrics, Behavior, and Identity Science, vol. 3, no. 2, pp. 229–239, 2021.

[8] P. H. Franses and T. Wiemann, "Intertemporal similarity of economic time series: An application of dynamic time warping," Computational Economics, vol. 56, no. 1, pp. 59–75, 2020.

[9] L. Nanni, "Experimental comparison of one-class classifiers for online signature verification," Neurocomputing, vol. 69, no. 7-9, pp. 869–873, 2006.

[10] S. S. Khan and A. Ahmad, "Relationship between variants of one-class nearest neighbours and creating their accurate ensembles," IEEE Transactions on Knowledge and Data Engineering, vol. 30, no. 9, pp. 1796–1809, 2018.

[11] H. Feng and C. C. Wah, "Online signature verification using a new extreme points warping technique," Pattern Recognition Letters, vol. 24, no. 16, pp. 2943–2951, 2003.

[12] M. Faundez-Zanuy, "Online signature recognition based on VQ-DTW," Pattern Recognition, vol. 40, no. 3, pp. 981–992, 2007.

[13] A. Parziale, M. Diaz, M. A. Ferrer, and A. Marcelli, "SM-DTW: Stability modulated dynamic time warping for signature verification," Pattern Recognition Letters, vol. 121, pp. 113–122, 2019.

[14] V. L. Blankers, C. E. van den Heuvel, K. Y. Franke, and L. G. Vuurpijl, "ICDAR 2009 signature verification competition," in 2009 10th International Conference on Document Analysis and Recognition, pp. 1403–1407, IEEE, 2009.

[15] M. I. Malik, M. Liwicki, L. Alewijnse, W. Ohyama, M. Blumenstein, and B. Found, "ICDAR 2013 competitions on signature verification and writer identification for on-and offline skilled forgeries (SigWiComp 2013)," in 2013 12th International Conference on Document Analysis and Recognition, pp.1477–1483, IEEE, 2013.

[16] Saleem, Mohammad, and Bence Kovari. "Preprocessing approaches in DTW based online signature verification." Pollack Periodica 15.1 (2020): 148-157.

[17] S. Lai and L. Jin, "Recurrent Adaptation Networks for Online Signature Verification," IEEE Trans. on Information Forensics and Security, vol. 14, no. 6, pp. 1624–1637, 2018.

[18] C. Li, X. Zhang, F. Lin, Z. Wang, J. Liu, R. Zhang and H. Wang, "A Stroke-based RNN for Writer-Independent Online Signature Verification," in Proc. International Conference on Document Analysis and Recognition (ICDAR), 2019.

**AUTHORS**

**Mohamad Saleem** is a Ph.D. candidate in software engineering at Budapest University of Technology and Economics, Hungary, at the Department of Automation and Applied Informatics. His research interests include online signature verification. He worked as a researcher and TA at Yarmouk University, Jordan, and Budapest University of Technology and Economics. He is a member of the Jordanian engineers' association.

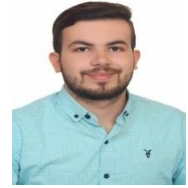

**Bence Kovari** received his Ph.D. degree in software engineering from Budapest University of Technology and Economics, Hungary, in 2013. Since then, he has been working as a researcher and teacher currently as an associate professor at the Department of Automation and Applied Informatics. His research interests include software engineering and automated verification of handwritten signatures. He has over 50 publications in the field. He is a member of the Hungarian Association for Image Processing and Pattern Recognition, a member of the John von Neumann Computer Society.

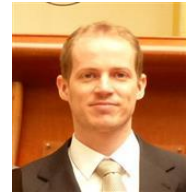